# LLMs' Understanding of Natural Language Revealed
(spoiler: LLMs **do not** understand language, but they can help us get there)

Walid S. Saba

*Institute for Experiential AI, Northeastern University, Portland, ME 04101*
w.saba@northeastern.edu

## Introduction

Large language models (LLMs) are the result of a massive experiment in bottom-up, data-driven reverse engineering of language at scale. Despite their utility in a number of downstream NLP tasks, ample research has shown that LLMs are incapable of performing reasoning in tasks that require quantification over and the manipulation of symbolic variables (e.g., planning and general problem solving) – see for example [25][26]. In this document, however, we will focus on testing LLMs for their language *understanding* capabilities, their supposed forte. In this regard we believe that we have not been testing the language **understanding** capabilities of large language models (LLMs) properly. Prompting LLMs and asking for responses will always look impressive because that's how LLMs were designed, i.e., to generate text. The proper method of testing the understanding capabilities of LLMs, we argue, is to prompt LLMs in reverse: give the LLM a snippet of text and query their understanding of the input text by asking the LLM questions against the input text.

As we will show here the language understanding capabilities of LLMs have been widely exaggerated. By testing the understanding capabilities properly – i.e., by giving the LLM snippets of text as input and then querying what the LLM "understood" it will become apparent that LLMs do not truly understand language, beyond very superficial inferences that are essentially the byproduct of the memorization of massive amounts of ingested text.

We have conducted here tests that involve the following linguistic phenomena:

(1) INTENSION
(2) KNOWLEDGE, BELIEF AND OTHER PROPOSITIONAL ATTITUDES
(3) COPREDICATION
(4) NOMINAL MODIFICATION
(5) METONYMY
(6) REFERENCE RESOLUTION (AND COMMONSENSE)

Additional tests are being conducted on the following:

(7) NOMINAL COMPOUNDS (OR, COMPOUND NOMINALS)
(8) DE RE / DE DICTO
(9) COMPOSITIONALITY
(10) PREPOSITIONAL PHRASE ATTACHMENTS
(11) QUANTIFIER SCOPE AMBIGUITIES

We will report on the tests (7) through (11) and update this document as the tests are completed.

# LLMs Do Not 'Understand' Language

Understanding a snippet of text involves, ultimately, building a mental picture or a mental model that corresponds to the state of affairs the text is supposedly describing. Moreover, "understanding" what was said also means making the right inferences from what was implicitly (if not explicitly) stated. We show here that LLMs do not truly 'understand' language and we do so by asking the LLM to describe various aspects of the given text. In querying the LLM we focus on a number of phenomena that have been extensively studied in linguistics, logic and the philosophy of language.

## INTENSION

Linguistic objects (words, phrases, sentences, mathematical expressions, etc.) have an extension and an intension (with an 's'). While the extension of a term or a linguistic object is its value (or what the object designates, or what it ultimately refers/evaluates to), the intension is the set of all properties that can be ascribed to the object. To take a simple example, consider the arithmetic expressions '2 * (3 + 4)' and '6 + 8'. While both expressions have the same extension (both evaluate to 14 – that is, both agree on the VALUE attribute), the two expressions have different intensions since there are several other properties that they do not share – e.g., NUM_OF_OPERATORS (that are 2 and 1, respectively) and NUM_OF_OPERANDS (that are 3 and 2, respectively)[1]. As another example consider '3 + 2' and 'number of UN countries with veto power'. While both expressions evaluate to 5, these two terms are not interchangeable in all linguistic contexts. To see this, consider the following:

(1) It will always be the case that [**MASK**] is 5

If [**MASK**] is replaced by '2 + 3' then (1) is always true, but if [**MASK**] is replaced by 'number of UN countries with veto power' then (1) is not necessarily true since we can imagine adding a country to or removing a country from the set of UN countries with veto power (note therefore that there can be a temporal aspect to intensions). From these examples it should be obvious that natural language is rampant with intensional contexts and this phenomenon has been studied – at least in a formal and mathematical sense, ever since Gottlob Frege's *Sense & Reference* [3] (see also [1][2][7] for introductory material and lecture notes on 'intension' and 'intensional semantics').

It should be noted, here, that the underlying architecture of LLMs – namely that of deep neural networks (DNNs) is purely extensional and thus these models cannot reason (cope) with intensionality. Below we give some examples that illustrates this very clearly.

---

**query**: Suppose "Billy the Kid is William H. Boney and that John loves to see movies about Billy the Kid." then is it true that "John loves to see movies about William H. Boney."?

**GPT4o**: Yes, in this case, it is true that "John loves to see movies about William H. Boney." Here's why: Since Billy the Kid and William H. Boney are the same person, movies about Billy the Kid are also movies about William H. Boney. Therefore, if John loves to see movies about Billy the Kid, he must also love to see movies about William H. Boney, as they are the same individual.

---

[1] Theoretically, the number of attributes an object has (or the number of ways one can describe an object) is infinite, although in a specific context there is small set of attributes that are considered relevant (see this for more details on why there's in theory an infinite number of attributes for any object).

**Problem:** Again the LLM made the wrong inference, due to the wrong 'understanding'. John might not know that Billy the Kid happens to be one 'William H Boney' and thus if John likes to see movies about Billy the Kid that does not mean John loves to see movies about William H Boney.

> **query** → Suppose "Mary told her friends that she enjoyed her vacation in Madrid, Spain." then is the following true: "Mary told her friends that she enjoyed her vacation in the Capital of Spain"?
>
> Yes, the statement "Mary told her friends that she enjoyed her vacation in the Capital of Spain" is true. Madrid is the capital of Spain, so both statements convey the same information about the location of Mary's vacation. ← **GPT4o**
>
> **2**

**Problem:** While the terms 'Madrid' and the 'Capital of Spain' are extensionally equal (they both refer to the same object, at least presently), their intensions are different and are therefore not interchangeable in every linguistic context since that would lead to the wrong conclusions: while Mary did tell her friends something about Madrid, she did not tell her friends anything about the Capital of Spain (like uncle Mitch, she might not even know that these two terms refer to the same object!)

> **query** → Suppose "Mary taught her little brother that 7 + 9 = 16." then is the following true: "Mary taught her little brother that 7 + 9 = SQRT(256)"?
>
> Yes, the statement "Mary taught her little brother that 7 + 9 = $\sqrt{256}$ is true. This is because $\sqrt{256}$ = 16 and the original statement confirms that Mary taught her little brother that 7 + 9 equals 16. Therefore, both statements are mathematically equivalent. ← **GPT4o**
>
> **3**

**Problem:** While $\sqrt{256}$ = 16 extensionally (they both refer to the same *value*) they are not equal intensionally and, again, are not interchangeable in every linguistic context. The inference above is wrong: if Mary taught her little brother that 7 + 9 = 16 this does not mean that she taught her little brother that 7 + 9 = $\sqrt{256}$. An intelligent NLU system should not make this inference as this would be the wrong understanding of what was said.

> **query** → Suppose "Mary loves visiting Paris, France." then is the following true: "Mary loves visiting the most populous city in France."?
>
> Yes, the statement "Mary loves visiting Paris, France" does imply that "Mary loves visiting the most populous city in France." Paris is the most populous city in France, so both statements convey the same information about the location that Mary loves to visit. ← **GPT4o**
>
> **4**

**Problem:** Similar to the previous example, while 'Paris' is in fact 'the most populous city in France' (both terms are extensionally equal as they both refer to the same object), these two terms have different intensions – for example, 'the most populous city in France' need not always be 'Paris' and so these two objects are not interchangeable in every linguistic context (uncle Mitch, for example, does not know that currently 'Paris' happened to be 'the most populous city in France').

> **prompt:** Suppose "John thinks that Aristotle was a brilliant philosopher." does that mean "John thinks that the tutor of Aleander the Great was a brilliant philosopher."?
>
> **GPT4o:** Yes, the statement "John thinks that Aristotle was a brilliant philosopher" does imply "John thinks that the tutor of Alexander the Great was a brilliant philosopher." This is because Aristotle is known to be the tutor of Alexander the Great.
>
> 5

**Problem:** Again, while 'Aristotle' and 'the tutor of Alexander the Great' refer to the same object, they are intensionally different and cannot be interchanged in every context (the understanding is wrong since John might not even know that 'the tutor of Alexander the Great' was 'Aristotle').

## NOMINAL MODIFICATION

Another linguistic phenomenon that has been extensively studied is that of nominal modification where there is usually one or more adjectives modifying a head noun (that could also be preceded by modifying nouns). There are several issues that arise in constructs such as [A+ N+], some of which can be illustrated by the following examples:

(1) a. Maria is an **ancient philosophy teacher**.
    b. Ron is an **admired philosophy teacher**.
(2) a. Olga is a **beautiful dancer**.
    b. Sara is a **tall dancer**.
(3) a. Susan made a **brilliant acquisition**.
    b. John made an **expensive acquisition**.

In (1a) it is clear that 'ancient' is modifying 'philosophy' while in (1b) 'admired' is modifying the head noun, 'teacher', since we do not usually speak of 'ancient teachers' but it is very common for us to speak of an 'admired teacher'. The issue in (2) is one of ambiguity, in that sometimes the noun modifier can be ambiguous as to what entity it is modifying. For example, while it is clear that 'tall' in (2b) is modifying Sara and not her dancing ('tall dancing'?), (2a) can be uttered by someone who thinks Olga is beautiful, or her dancing (or both!). Similarly, in (3b) 'expensive' is clearly modifying the acquisition, not John, while in (3a) 'brilliant' must be modifying Susan (it was brilliant of Susan to make the acquisition). Clearly, all of these interpretations are a function of commonsense knowledge of how we talk about entities in our everyday spoken language. For introductory material on 'nominal modification' see [13] and for more advanced studies on the role nominal modification plays in the semantics of natural language see [12][14]. We now go through some examples involving nominal modification and show that LLMs also fail to capture the real semantic content.

> **query:** Show in Prolog code the entities with their types and the relations that are implicit in the following text: "Sara is an experienced and talented dancer."
>
> **GPT4o:**
> ```
> % Entities with their types
> entity(sara, person).
> entity(dancer, profession).
> % Relations
> relation(is_a, sara, dancer).
> attribute(sara, experienced).
> attribute(sara, talented).
> ```
> 6

**Problem:** Here's a simple explanation of why the LLM made the wrong conclusion as what the text means. If in some other snippet of text we stated that (the same) Sara is a teacher, then with the above understanding we will easily (and wrongly) infer (by conjunction) that Sara is an experienced and talented teacher, which is a wrong inference. The point here is that the above text does not state that Sara is experienced and talented (as a person), but only as a dancer.

---

**query** → Show in Prolog code the entities with their types and the relations that are implicit in the following text: "Mary appreciated the generous gift John gave her. It was a beautiful watch."

← GPT4o

```
% Entities and their types
entity(mary, person).
entity(john, person).
entity(gift, object).
entity(watch, object).
% Relations
relation(appreciated, mary, gift).
relation(gave, john, gift).
relation(recipient, mary, gift).
attribute(gift, generous).
relation(is_a, gift, watch).
attribute(watch, beautiful).
```

---

**Problem:** The LLM also failed our last example on nominal modification by suggesting that a gift, which is an inanimate object (a watch), is what is generous, although what the text implies is that 'it was generous of John to give Mary the watch as a gift'!

---

**query** → Show in Prolog code the entities with their types and the relations that are implicit in the following text: "Susan made a brilliant and timely acquisition."

← GPT4o

```
% Entities with their types
entity(e1, person, 'Susan').
entity(e2, adjective, 'Brilliant').
entity(e3, adjective, 'Timely').
entity(e4, action, 'Acquisition').
% Relations
performed(e1, e4).         % Susan performed the Acquisition
has_attribute(e4, e2).     % Acquisition has the attribute Brilliant
has_attribute(e4, e3).     % Acquisition has the attribute Timely
```

---

**Problem:** Since 'brilliant' syntactically modifies 'acquisition' the LLM wrongly inferred that 'brilliant' is modifying the 'acquisition' action (or activity). Note that it correctly (although for the wrong reason) inferred that 'timely' modifies 'acquisition' it could not infer that something that could be 'timely' could not be 'brilliant' – an entity is either a person (that could be 'brilliant') or an activity, that could be 'timely', but not both!

Again we should note that these seemingly simple mishaps in understanding the text are not trivial since we are ultimately interested in building formal structures (e.g., knowledge graphs) from raw text in such a way that our subsequent queries and inferences would result in correct answers.

> **query** → Show in Prolog code the entities with their types and the relations that are implicit in the following text: "Ron is an expert in classical Italian music."
>
> **% Entities with their types**
> entity(ron, person).
> entity(music, field).
> entity(classical, property).
> entity(italian, property).
> **% Relations**
> relation(expert_in, ron, music).
> attribute(music, classical).
> attribute(music, italian).
>
> ← GPT4o
>
> 9

**Problem:** Again, the LLM made the wrong inference here since the text states that Ron is an expert in classical **Italian** music and not in classical music in general! Again, while this error might seem trivial, when combined with other errors on other snippets of text, the combined result would be a complete misunderstanding of the larger piece of text.

> **query** → Show in Prolog code the entities with their types and the relations that are implicit in the following text: "Australia beat the Canadian soccer team in the World Cup 3-2."
>
> **% Entities with their types**
> entity(australia, country).
> entity(canadian_soccer_team, team).
> entity(world_cup, event). entity(score, result).
> **% Relations**
> relation(beat, australia, canadian_soccer_team).
> relation(in_event, beat, world_cup).
> relation(score_of, beat, score).
> value(score, 3-0).
>
> ← GPT4o
>
> 10

**Problem:** In addition to several problems in the interpretation, the LLM wrongly inferred that Australia the country (which is a geographic location) and not the Australian soccer team is the one that beat the Canadian soccer team. Commonsense tell us, however, that what have here is metonymy where Australia is used to refer to another entity that Australia is related to, and in particular, the Australian national soccer team.

> **query** → Show in Prolog code the entities with their types and the relations that are implicit in the following text: "Ron is a formal philosophy teacher."
>
> **% Entities with their types**
> entity(ron, person).
> entity(philosophy_teacher, profession).
> **% Relations**
> relation(is_a, ron, philosophy_teacher).
> attribute(ron, formal).
>
> ← GPT4o
>
> 11

**Problem:** The LLM fails to infer that 'formal' modifies 'philosophy' and that 'Ron is a formal philosophy teacher' should be understood as [Ron is a [formal philosophy] teacher] and that Ron is not 'a formal' – not even a 'formal teacher'!

## KNOWLEDGE, BELIEF AND OTHER PROPOSITIONAL ATTITUDES

Propositional attitudes are mental states held by some agent about some proposition (statement). Consider for example the following proposition (statement):

(2) $P$ = The Statue of Liberty was a gift from the people of France to the United States.

An agent Jon could have several propositional attitudes towards the proposition in (1) such as:

(3) **knows**(Jon, $P$)   Jon knows [that] $P$
(4) **believes**(Jon, $P$)   Jon believes [that] $P$
(5) **thinks**(Jon, $P$)   Jon thinks [that] $P$

Note that propositional attitudes are (usually) linguistically marked by some verb such as *thinks*, *knows*, *believes*, *wants*, *intends*, *desires*, *wishes*, *hopes*, etc. that some proposition $P$ (is true)[2]. Note also that there's a difference between knowledge, belief, and truth. That is, some intelligent agent A might know that some proposition $P$ (is true) but they might also just 'think' (or 'believe') that some proposition $P$ (is true) where in fact the proposition $P$ is not true (if "Jon thinks/believes [**the earth is flat**]" that does not make [**the earth is flat**] true)[3]. (see [8][9] for introductory material on 'propositional attitude' and [10][11] for more formal discussions)

What is important for us here is to investigate how much do LLMs 'understand' the subtle difference between truth (facts), knowledge and belief in ordinary spoken language. Below we show that in most cases LLMs fail to make the right inference in contexts with propositional attitudes.

> **query:** Suppose "Stavros, who was just elected as the mayor of Athens, and John know each other very well as they became friends when they both attended the same university in the UK." then does John know any mayors?
>
> **GPT4o:** Yes, based on the information provided, John does know at least one mayor. Stavros, who has just been elected as the mayor of Athens, knows John very well as they became friends when they both attended the same university in the UK. Therefore, John knows Stavros, who is a mayor.
>
> 12

**Problem:** The 'understanding' of the LLM is wrong. While John knows Stavros well, this does not mean he knows every true fact about Stavros (e.g., that Stavros was elected the mayor of Athens).

---

[2] Note that the relationship between an agent and propositional attitudes is man-to-many: an agent might have several propositional attitudes towards a specific proposition; and an agent could also have the same propositional attitude towards several propositions.
[3] Some philosophers (of language and mind) have tried to formalize the relationship between various propositional attitudes. For example, Fodor [5] formalizes the relationship between 'want' and 'belief' as it relates to the psychology of desires as such: "If $x$ **wants** that $P$, and x **believes** that not-$P$ unless $Q$, and x believes that $x$ can bring it about that $Q$, then $x$ tries to bring it about that $Q$".

> **query:** Suppose "Sara thinks everything Galileo believes is true, and Galileo believes that Mars rotates on its axis, which is actually true." does that mean "Sara knows that the red planet rotates on its axis."?
>
> **GPT4o:** Assuming that Sara has good reasons to consider Galileo a credible and reliable authority on astronomy, her belief in his claim about Mars would be justified. In this case, we can conclude that Sara's belief is a justified true belief, which means she knows that the red planet rotates on its axis.
>
> 13

**Problem:** There are several problems here. First, Sara might not know that Mars is referred to by 'the red planet' and that alone should block the inference that Sara knows (or even believes) that Mars rotates on its axis. Second, the LLM failed to recognize the difference between 'believes' and 'knows'. If Sara believes everything Galileo says, then at the most she can believe that Mars rotates on its axis, but we cannot say she 'knows' that. For Sara to 'know' something (and not merely believe it), she must be aware of the truth of that fact, and nothing in the text implies that.

> **query:** Suppose that "On his trip to Paris, Carlos met Sophia, a beautiful lady who is originally a Romanian Gypsy. They had a great week and became very good friends, but like everything in life, everything has an end." then does Carlos know any gypsy lady?
>
> **GPT4o:** Yes, based on the information provided, Carlos does know a gypsy lady. Sophia, whom he met on his trip to Paris and became very good friends with, is originally a Romanian Gypsy. Therefore, Carlos knows Sophia, who is a gypsy lady.
>
> 14

**Problem:** While Carlos knows Sofia, that does not mean Carlos knows every true fact about Sofia (such as the fact that she is a Romanian Gypsy).

> **query:** Suppose "John knows Maria who is a teacher, that also happens to be a talented dancer." Then does John know a dancer?
>
> **GPT4o:** Since John knows Maria and Maria is a talented dancer, it follows logically that John knows at least one talented dancer, which is Maria. Therefore, **yes, John knows a dancer**.
>
> 15

**Problem:** Again, the LLM arrived at the wrong conclusion (understanding): While John knows Maria (who happens to also be a talented dancer), it was never stated that John also knows she is a talented dancer so we cannot conclude that John (is aware that he actually) knows a dancer!

> **query:** Suppose that "John knows that interns at the company must be graduate students and suppose that Mary is an intern at the company" then does John know that Mary is a graduate student?
>
> **GPT4o:** Yes. Since John knows that "interns at the company must be graduate students" and Mary being an intern falls under this rule, John can logically deduce that Mary is a graduate student based on his knowledge of the rule. Therefore, we can conclude that John knows that Mary is a graduate student.
>
> 16

**Problem:** Again, the LLM made the wrong inference here (wrong 'understanding'). While John knows the rules about interns, and while Mary is an intern, John does not know that she is (this was never stated), and so he cannot infer anything about Mary. For the rule of interns to apply (in John's head), he has to have access to the knowledge that Mary is an intern!

---

**query:** Suppose "John knows Clara who is a cellist and suppose also that Clara is recognized as a talented teacher." then if we ask John whether he knows someone who is talented, what should he answer?

**GPT4o:** Based on the information provided, John knows Clara, who is recognized as a talented teacher. Given this context, if John is asked whether he knows someone who is talented, he should answer "Yes."

17

---

**Problem:** The LLM made the wrong inference here also since John, while he knows Clara as a person, and even if he knew that she is a cellist, he does not necessarily know some fact about Clara, specifically that she happens to also be a talented teacher, and so he should not confirm that he does know any talented teachers.

## COPREDICATION

Copredication is the phenomenon that two or more predicates (properties, or relations) are being applied to the same entity or reference thus making a single reference being used to refer – at once, to several entities of different types (see [4][6]). Recognizing all the different types of entities being implied by a single reference requires deep understanding of the overall context. As we show here, LLMs also fail to recognize copredication, even in its simplest form.

---

**query:** Show me, in Prolog code, all the entities with their types as well as the primitive relations that are implicit in this text, giving objects unique IDs for further reference: "I was in the city of Barcelona when the city was celebrating its win over Real Madrid as it was also getting to ready to vote for independence."

**GPT4o:**
```
% Entities with their types
entity(e1, person, 'I').
entity(e2, city, 'Barcelona').
entity(e3, soccer_team, 'Real Madrid').
entity(e4, event, 'Win').
entity(e5, event, 'Vote for Independence').
% Relations
in(e1, e2).                    % 'I' was in 'Barcelona'
celebrating(e2, e4, e3).       % 'Barcelona' was celebrating its 'Win' over 'Real Madrid'
preparing(e2, e5).             % 'Barcelona' was getting ready to 'Vote for Independence'
```

18

---

**Problem:** The LLM did not recognize the copredication of 'Barcelona' where the entity is being used in three senses at the same time: (i) as the visited **city**, (ii) as the **football team** that won over Real Madrid, and (iii) as the **citizens of the city** that were celebrating and that will be voting for independence. The LLM wrongly inferred that the city (the geographic location) is what won over Real Madrid, and that the geographic location is what celebrated and will be voting, etc.

**query** → Show me the basic entities (with their types and IDs) and the basic relations between them as implied by the following text: "I threw away the newspaper I was reading as I knew it fired my favorite columnist."

GPT4o

```
% Entities with their types
entity(e1, person, 'I').
entity(e2, object, 'Newspaper').
entity(e3, person, 'Favorite Columnist').
entity(e4, event, 'Firing').
% Relations
reading(e1, e2).           % 'I' was reading the 'Newspaper'
threw_away(e1, e2).        % 'I' threw away the 'Newspaper'
knew_about(e1, e4).        % 'I' knew about the 'Firing'
involved(e4, e3).          % 'Firing' involved the 'Favorite Columnist'
caused(e2, e4).            % The 'Newspaper' caused the 'Firing'
```

19

**Problem:** Again, the LLM did not recognize the copredication in the text, namely that the physical newspaper object (that was thrown away) is not who fired 'my favorite columnist': the firing was done by the [editorial board/management] of the newspaper and not by my physical newspaper.

**query** → Show me, in Prolog code, all the entities with their types as well as the primitive relations that are implicit in this text, giving objects unique IDs for further reference: "I bought Das Kapital by Karl Marx, a book that influenced many revolutions."

GPT4o

```
% Entities with their types
entity(e1, person, 'I').
entity(e2, book, 'Das Kapital').
entity(e3, author, 'Karl Marx').
entity(e4, concept, 'Revolutions').
% Relations
bought(e1, e2). % 'I' bought 'Das Kapital'
written_by(e2, e3). % 'Das Kapital' was written by 'Karl Marx'
influenced(e2, e4). % 'Das Kapital' influenced 'Revolutions'
```

20

**Problem:** Again, the LLM did not recognize the copredication in the text – the LLM inferred that the physical object that I bought (the actual book that I own), and not the content and ideas in all copies of Das Kapital, is what influenced many revolutions, which is obviously wrong.

**query** → Show me, in Prolog code, all the entities with their types as well as the primitive relations that are implicit in this text: "I was partying in Russia when it started moving into Ukraine."

GPT4o

```
% Entities with their types
entity(e1, person, 'I').
entity(e2, country, 'Russia').
entity(e3, country, 'Ukraine').
entity(e4, activity, 'Partying').
entity(e5, military_action, 'Moving into').
% Relations
was_partying_in(e1, e2, e4). % 'I' was partying in 'Russia'
initiated_action(e2, e5, e3). % 'Russia' started moving into 'Ukraine'
```

21

Walid S. Saba

**Problem:** Again, the LLM did not recognize the copredication in the text – the LLM inferred that the country I was visiting (and partying in) and not its army is what started moving into Ukraine.

It should be noted here that we are aware of the fact that some readers might not (fully) appreciate the point of highlighting these subtle errors in understanding. What we can briefly say here is this: undoubtedly, we expect an intelligent NLU system that reads some text to convert the informal text into some formal structure that can subsequently be queried. In the above example, the system must therefore decide on the entity that 'Russia' refers to: is it a geographic location (that surely cannot move into another country) or does it also (implicitly) refer to some other entity that is related to Russia (such as the Russian army). Our aim therefore is not to dwell on these subtle failures in understanding of the text, but to highlight the fact that building an AI that fully understands natural language text is not as simple as most superficial studies have, unfortunately, concluded.

Let us now conclude testing LLMs understanding of copredication with one final example.

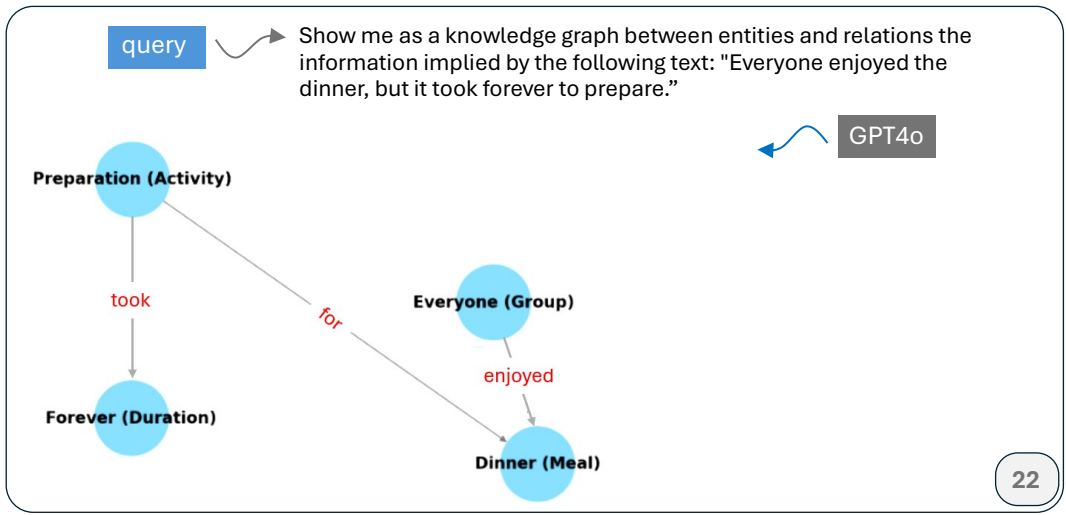

**Problem:** Again, the LLM did not recognize the copredication in the text – the LLM inferred that the dinner **event** (that everyone enjoyed) is what "took forever to prepare" while what took a long time to prepare is the meal and not the (dinner) event. That is, it was the food of the dinner that took a long time to prepare, and not the sitting during the dinner that everyone enjoyed[4].

## METONYMY

Metonymy occurs when we use an entity e1 to refer indirectly to another entity e2 that stands in some relation to e1. In the course of 'understanding' we usually resort to our commonsense background knowledge to infer the implicit relation between e1 and e2 (see [16][17] and [18]).

---

[4] Querying what the LLM understood can be done by asking for the entities and relations in code (such as Prolog or RDF) or in graphical form as a knowledge graph (or an ER diagram)

> **query** Show in Prolog code the entities with their types and the relations that are implicit in the following text: "I was reading the New York Times that I borrowed from John, and it seems that they will be endorsing Joe Biden for president."
>
> **% Entities and their types**
> entity(i, person).
> entity(john, person).
> entity(new_york_times, publication).
> entity(joe_biden, person).
> entity(president, position).
> **% Relations**
> relation(borrowed_from, i, john, new_york_times).
> relation(reading, i, new_york_times).
> relation(will_endorse, new_york_times, joe_biden).
> relation(for_position, joe_biden, president).
>
> GPT4o
>
> 23

**Problem:** According to the LLM an object that I borrowed from John (namely the physical copy of the New York Times that I was reading) is endorsing Joe Biden (what would happen if I destroyed the physical copy I was reading? Will the endorsement disappear?) Clearly, the endorsement of Joe Biden was made by New York Times editorial board and not by the physical copy of the newspaper that is in my hands.

> **query** Show in Prolog code the entities with their types and the relations that are implicit in the following text: "As I entered the restaurant, I heard the waiter say that the loud omelet wants another beer."
>
> **% Entities and their types**
> entity(i, person).
> entity(restaurant, location).
> entity(waiter, person).
> entity(omelet, food).
> entity(beer, drink).
> **% Relations**
> relation(entered, i, restaurant).
> relation(heard, i, waiter).
> relation(say, waiter, omelet_wants_beer).
> relation(wants, omelet, beer).
> attribute(omelet, loud).
> quantity(beer, another).
>
> GPT4o
>
> 24

**Problem:** While there are several failures here (e.g., that 'omelet wants a beer' is interpreted as a relation between a 'say' and a 'waiter', the main failure here is that 'loud omelet' is taken literally, while 'the loud omelet' is a reference here to 'the loud [**person eating the**] omelet'. These metonymic contexts are usually understood by resorting to commonsense knowledge ([14] gives a detailed explanation of how this happens using ontological types and type unification).

> **query** → Show in Prolog code the entities with their types and the relations that are implicit in the following text: "Australia beat the Canadian soccer team in the World Cup 3-2."
>
> **% Entities with their types**
> entity(australia, country).
> entity(canadian_soccer_team, team).
> entity(world_cup, event). entity(score, result).
> **% Relations**
> relation(beat, australia, canadian_soccer_team).
> relation(in_event, beat, world_cup).
> relation(score_of, beat, score).
> value(score, 3-0).
>
> ← GPT4o
>
> 25

**Problem:** In addition to several problems in the interpretation, the LLM wrongly inferred that Australia the country (which is a geographic location) and not the Australian soccer team is the one that beat the Canadian soccer team. Commonsense tell us, however, that what have here is metonymy where Australia is used to refer to another entity that Australia is related to, and in particular, the Australian national soccer team.

> **query** → Show in Prolog code the entities with their types and the relations that are implicit in the following text: "The taxi I took to get here was very friendly."
>
> **% Entities with their types**
> entity(taxi, vehicle).
> entity(i, person).
> entity(here, location).
> **% Relations**
> relation(took, i, taxi).
> relation(to_get_to, i, here).
> attribute(taxi, friendly).
>
> ← GPT4o
>
> 26

**Problem:** The final example involving metonymy shows also how LLMs fail to 'infer' the implicit commonsense relationships that we implicitly assume in our ordinary spoken language. While the LLM inferred that 'the taxi I took' refers to a vehicle, it did not infer that referring to the taxi as being friendly is a reference to the driver of the taxi and not the vehicle. Note that these are not exotic or farfetched examples– this in fact is how we mostly speak, and thus an AI that truly understands natural language must make these simple inferences.

## REFERENCE RESOLUTION AND COMMONSENSE

Reference resolution is perhaps the most known phenomenon in natural language semantics. A reference is usually a pronoun (or relative pronoun) that refers to some entity that is mentioned elsewhere in the wider context. Note that the general utility of 'referring' is called 'anaphora' of which pronouns are the most commonly used tool. (see [19][20][21] and [22]).

> **query**: What does 'she' refer to in the following: "Mary asked Sara when does the library closes since she always geos there."
>
> **GPT4o**: In the sentence "Mary asked Sara when does the library close since she always goes there," the pronoun "she" is ambiguous and could refer to either Mary or Sara. Without additional context, it is not possible to definitively determine which person "she" refers to.
>
> 27

**Problem:** While, in theory, 'she' in the above could refer to either Mary or Sara, commonsense tells us that the ones that always goes to the library would know when the library closes more than others. Moreover, why would Mary ask Sara when the library closes if it was Mary that always geos there?

> **query**: What does 'he' refer to in the following: "a teenager shot a policeman, but eyewitnesses said he managed to flee away and escape fatal injuries."
>
> **GPT4o**: In the sentence "A teenager shot a policeman, but eyewitnesses said he managed to flee away and escape fatal injuries," the pronoun "he" most likely refers to the teenager. The structure of the sentence and typical narrative focus suggest that the eyewitnesses' statement pertains to the person who performed the significant action (the shooting), which is the teenager in this case.
>
> 28

**Problem:** While the syntactic structure might favor 'he' referring to the teenager, commonsense says that the one trying to escape further injuries is the one that was shot, namely the policeman. Since LLMs learned statistical correlations, it could also be the that the wrong inference was made here since the correlation between someone shooting and someone escaping should be high.

> **query**: What does 'he' refer to in the following: "Dr. Smith told John that he should be done reading his thesis next week."
>
> **GPT4o**: In the sentence "Dr. Smith told John that he should be done reading his thesis next week," the pronoun "he" is ambiguous and could refer to either Dr. Smith or John. Without additional context, it is not possible to definitively determine which person "he" refers to.
>
> 29

**Problem:** Again, while the syntactic structure permits both possible resolutions of 'he', commonsense should favor 'he' referring to Dr. Smith since John, who wrote the thesis has surely read his own writing, while it is Dr. Smith that perhaps did not yet read the thesis that John wrote!

> **query**: What does 'she' refer to in the following: "Mary asked Sara what does she need to bring to the party."
>
> **GPT4o**: In the sentence "Mary asked Sara what does she need to bring to the party," the pronoun "she" most likely refers to Sara. The structure of the sentence suggests that Mary is inquiring about what Sara needs to bring to the party. Thus, "she" refers to Sara.
>
> 30

**Problem:** The LLM made the wrong inference here because we do not usually ask someone else what they **need** to bring to a party, but we ask what do we need to bring. The inference would have been correct if the input text was changed to "Mary asked Sara what is she going to bring to the party." – since we don't ask someone about what we are going to bring, but perhaps about what they are going to bring.

> query → Mary and John took the kids in their class to the zoo. They spent most of the day asking about the diet different animals require." then what does "they" refer to?
>
> In the sentence, "Mary and John took the kids in their class to the zoo. They spent most of the day asking about the diet different animals require," the pronoun "they" is ambiguous. It could refer to either "Mary and John" or "the kids in their class." The context does not provide enough information to definitively determine which group "they" refers to. ← GPT4o
>
> 31

**Problem:** The LLM made the wrong inference here: it is much more likely that the students were asking their teachers questions about the animals then the other way around and so by relying on our commonsense there should not be any ambiguity.

# LLMs Can Help Us Get There

To be written.

# Concluding Remarks

To be written.